%% file: main.tex
\documentclass[preprint,12pt]{elsarticle}

\usepackage{ifpdf}
\usepackage{graphicx}

\usepackage{amsmath}
\usepackage{algorithmic}
\usepackage{array}

\usepackage{amsfonts}
\usepackage{amssymb}
\usepackage{float}
\usepackage{hyperref}
\usepackage{color}
\usepackage{graphicx}
\usepackage{caption}
\usepackage{subcaption}
\usepackage[utf8]{inputenc}
\usepackage{url} 
\usepackage{mathtools}
\usepackage{amsmath}
\usepackage{geometry}
\usepackage{bm}

\title{Complex Recurrent Spectral Network}

\begin{document}
\begin{frontmatter}

\author[1,*]{Lorenzo Chicchi}
\author[1,2]{Lorenzo Giambagli}
\author[1]{Lorenzo Buffoni}
\author[1]{Raffaele Marino}
\author[1]{Duccio Fanelli}

\address[1]{Department of Physics and Astronomy, University of Florence, INFN and CSDC, Sesto Fiorentino, Italy}
\address[2]{naXys - Namur Center for Complex Systems, University of Namur, rue Graf\'e 2, 5000 Namur, Belgium}
\address[*]{Corresponding Author: lorenzo.chicchi@unifi.it}

\end{frontmatter}

\input{ComplexRSN}

\section*{Acknowledgments}

This work is supported by \#NEXTGENERATIONEU (NGEU) and funded by the Ministry of University and Research (MUR), National Recovery and Resilience Plan (NRRP), project MNESYS (PE0000006) "A Multiscale integrated approach to the study of the nervous system in health and disease" (DN. 1553 11.10.2022).

\bibliographystyle{unsrt}
\bibliography{references.bib}

\end{document}

%% file: ComplexRSN.tex
\newcommand{\xnl}[0]{ \vec{x}^{\hspace{1mm}nl}}
\newcommand{\xl}[0]{ \vec{x}^{\hspace{1mm}l}}
\newcommand{\ynl}[0]{ \vec{y}^{\hspace{1mm}nl}}
\newcommand{\yl}[0]{ \vec{y}^{\hspace{1mm}l}}
\newcommand{\wnl}[0]{ \vec{w}^{\hspace{1mm}nl}}
\newcommand{\wl}[0]{ \vec{w}^{\hspace{1mm}l}}
\newcommand{\crsn}[0]{$\mathbb{C}$-RSN}

\section*{Abstract}

This paper presents a novel approach to advancing artificial intelligence (AI) through the development of the Complex Recurrent Spectral Network ($\mathbb{C}$-RSN), an innovative variant of the Recurrent Spectral Network (RSN) model. The $\mathbb{C}$-RSN is designed to address a critical limitation in existing neural network models: their inability to emulate the complex processes of biological neural networks dynamically and accurately. By integrating key concepts from dynamical systems theory and leveraging principles from statistical mechanics, the $\mathbb{C}$-RSN model introduces localized non-linearity, complex fixed eigenvalues, and a distinct separation of memory and input processing functionalities. These features collectively enable the $\mathbb{C}$-RSN evolving towards a dynamic, oscillating final state that more closely mirrors biological cognition. Central to this work is the exploration of how the $\mathbb{C}$-RSN manages to capture the rhythmic, oscillatory dynamics intrinsic to biological systems, thanks to its complex eigenvalue structure and the innovative segregation of its linear and non-linear components. The model's ability to classify data through a time-dependent function, and the localization of information processing, is demonstrated with an empirical evaluation using the MNIST dataset. Remarkably, distinct items  supplied as a sequential input yield patterns in time which bear the indirect imprint of the insertion order (and of the time of separation between contiguous insertions).  

\section{Introduction}

Today, artificial intelligence (AI) \cite{XU2021100179} stands as a cornerstone of innovation and progress. The surge in AI research and its applications across various domains, from healthcare \cite{esteva2019guide, ching2018opportunities} and finance \cite{sezer2020financial,marino2023solving} to autonomous systems \cite{he2018amc, grigorescu2020survey} and beyond \cite{marino2023hard,marino2023solving,marino2021learning,preChicchi,chen2018neural}, underscores its growing importance. At the heart of this burgeoning field is the quest to not only replicate but also to enhance human cognitive abilities through computational means \cite{roman2017cognitive, kriegeskorte2018cognitive}. AI’s significance is particularly evident in its ability to process vast amounts of data \cite{CHICCHI2023118352}, uncover patterns \cite{bishop2006pattern,shalev2014understanding}, and make decisions \cite{marino2016backtracking}, often surpassing human capabilities in speed and efficiency \cite{silver2016mastering}.

Central to the evolution of AI are neural networks \cite{prince2023understanding}, which draw inspiration from the biological neural networks \cite{kriegeskorte2015deep} that constitute animal brains. These networks are systems of interconnected nodes, or neurons, that work in concert to solve complex problems. Among these, Recurrent Neural Networks (RNNs) \cite{lecun2015deep, goodfellow2016deep, chicchi2023recurrent} have emerged as a pivotal tool, especially in tasks involving sequential data. Unlike traditional neural networks, RNNs possess a unique feature: memory. This allows them to process not just individual data points, but entire sequences of data, making them particularly adept at tasks like language processing \cite{collobert2008unified}, time series analysis \cite{selvin2017stock}, and even music composition \cite{eck2002first,agarwala2017music}.

Despite the impressive performance displayed by AI, ML, and DL technplogies, a significant gap remains in our understanding of their inner workings \cite{zhang2017mdnet}. This opacity often limits their broader application, especially in scenarios demanding transparency and reliability \cite{eshete2021making, rudin2019stop}. Indeed, interpretability \cite{linardatos2020explainable, conmy2023towards} represents still a critical challenge in the field.

Physicists have begun to elaborate on these aspects, by employing concepts from Statistical Mechanics \cite{huang2009introduction, baldovin2023ergodic, marino2016advective, marino2016entropy,pittorino2017chaos, aurell2016diffusion, caracciolo2021criticality} to construct theoretical frameworks for AI systems \cite{baldassi2021unveiling,baldassi2022learning}. Their approaches typically utilize models of disordered \cite{lucibello2022deep, pacelli2023statistical, agliari2023parallel, marino2023phase,angelini2023stochastic} or complex systems \cite{giambagli2021machine, marino2023large, buffoni2022spectral, chicchi2021training,chicchi2023recurrent}, shedding light on the intricate dynamics of AI algorithms. In this paper we adopt a distinct perspective, by leveraging the principles of dynamical systems theory \cite{vulpiani2009chaos,strogatz2018nonlinear,ott2002chaos,weinan2017proposal}. This approach is particularly suited to unravel the complexities of AI, and thus offering a more structured and theoretically grounded understanding to their inherent functioning.

The recent introduction of EODECA (Engineered Ordinary Differential Equation as Classification Algorithms) \cite{marino2023bridge} in the literature marks a significant leap forward in this endeavor. EODECA establishes a novel connection between machine learning and dynamical systems, by opening up the perspective for a fresh lens through which AI algorithms can be examined and understood. 

In this paper, our focus is to delve deeper into the intricacies of Recurrent Neural Networks (RNNs). More specifically, by applying the principles of dynamical systems theory, we aim to enhance and expand upon the existing work on Recurrent Spectral Methods (RSN), as introduced in \cite{chicchi2023recurrent}. In this latter paper,  a novel machine learning strategy was proposed wherein the evaluation process is intricately linked to the dynamics of a specifically engineered system. The spectral parametrization of the adjacency matrix within a fully connected network. allows to effectively steer the system's dynamics towards a pre-defined vector subspace, spanned by select eigenvectors. During the learning phase, parameters are self-consistently chosen so as to ensure that the system converges to a final stationary state, which points to the class the provided input data belongs to. This methodology has revealed several notable properties. A striking observation is the apparent independence of the evaluation process from the number of steps, or the system's integration time, as employed during training. This results in a robust convergence of accuracy (or measured loss) to a stable asymptotic value, which remains consistent across successive iterations, transcending the constraints of the training's temporal horizon. This characteristic markedly contrasts with traditional Recurrent Neural Networks (RNNs), where the iteration count is a critical parameter for accurately categorizing test data.

However, despite these advancements, the RSN model exhibits a significant limitation: it predicts a static final state. This stands in stark contrast to the dynamic nature of real biological systems, where neurons exhibit continuous, time-evolving activity. 

To address this critical shortcoming, this manuscript introduces an innovative variant of the RSN model: the Complex Recurrent Spectral Network $(\mathbb{C}$-RSN). The $\mathbb{C}$-RSN model endeavors to move beyond the static final state paradigm of the standard RSN model. Through this enhancement, $\mathbb{C}$-RSN aims to offer a more accurate and biologically-relevant representation of neural processes, potentially unlocking new horizons in the application of machine learning to complex, time-sensitive tasks.

Building on the foundation of the RSN model, the Complex Recurrent Spectral Network ($\mathbb{C}$-RSN) introduces several key innovations that significantly enhance its capability to model dynamic systems, drawing it closer to the complexity of biological neural networks. These new elements are intricately woven into the fabric of the $\mathbb{C}$-RSN model, each serving a distinct and critical role in the overall functionality:
\begin{itemize}
    \item \textbf{Localized Non-linearity in a Subset of Nodes}: In the $\mathbb{C}$-RSN model, non-linearity is not uniformly distributed across the entire network. Instead, it is strategically localized within a specific subset of nodes, referred to as the \textit{non-linear part}. This targeted application of non-linearity allows for a more nuanced and controlled interaction within the network.
    \item \textbf{Complex Fixed Eigenvalues}: A fundamental aspect of the $\mathbb{C}$-RSN is the incorporation of complex fixed eigenvalues of the form $e^{2\pi i /T_m}$. This complex eigenvalue structure imbues the network with a rhythmic, oscillatory dynamic that is reminiscent of the cyclical processes observed in biological neural activities \cite{ tsuda2001toward, chalk2016neural}. This feature allows the network to model time-dependent phenomena more effectively, providing a richer, more versatile framework for understanding temporal patterns \cite{ scott1981patterns}.
    \item \textbf{Memory Confined in the Linear Part}:In the $\mathbb{C}$-RSN architecture, memory functionality is confined to the linear part of the network. This segregation of memory into a dedicated linear subsystem ensures a more stable and reliable retention of information. The system is also able to handle sequential inputs by keeping track of the relative insertion order.
    \item \textbf{Input Processing in the Non-Linear Part}: The design of the $\mathbb{C}$-RSN ensures that the input is primarily processed in the non-linear part of the network. This approach allows for a more dynamic and adaptive response to incoming data, as the non-linear elements of the network provide a rich, flexible mechanism for data interpretation and response. 
 \end{itemize}

These innovative features collectively empower the $\mathbb{C}$-RSN model to transcend the limitations of the static RSN framework, offering a more dynamic, adaptive, and biologically-relevant system. The integration of localized nonlinearity, complex eigenvalues, dedicated memory storage, and specialized input processing marks a significant step forward in the development of neural network models that more accurately reflect the intricate and dynamic nature of biological neural processes.

The introduction of these novel components into the $\mathbb{C}$-RSN model heralds a paradigm shift in the way the system approaches its final state. As we delve deeper in the subsequent sections, it becomes evident that the system no longer converges to a static state. Instead, it evolves towards a dynamic, oscillating final state. This behavior is a direct consequence of the innovative elements integrated into the $\mathbb{C}$-RSN model.

The oscillatory nature of the final state is intricately linked to the complex and fixed eigenvalues, specifically those of the form $e^{2\pi i /T_m}$. These eigenvalues play a crucial role in defining the temporal dynamics of the system. The final state emerges as a linear superposition of eigenvectors associated with these eigenvalues, leading to a periodic behavior where the period is contingent upon the values of the constants $T_m$. This introduces a new dimension of temporal dynamics to the model, allowing it to simulate the rhythmic and oscillatory processes characteristic of biological systems.

Furthermore, the emergent wavefront's shape in this final state reflects the specific combination of eigenvectors that remain active in the asymptotic state. In essence, the complex eigenvalues form a foundational basis of frequencies or periods, providing a framework upon which the state can be articulated at large timescales. This ability to represent the system's state as a function of time through a combination of frequencies is a significant advancement, offering a novel approach to understanding and modeling dynamic processes.

In practical terms, this allows the $\mathbb{C}$-RSN to address classification problems in a unique manner. By initiating the network with an activity corresponding to a specific dataset, the network is tasked with generating a time-dependent function $f_C(t)$, which is defined by the activity within a sub-portion of the network. This function effectively becomes a temporal signature of the classified data. $\mathbb{C}$-RSN possesses the capability to sequentially process multiple sets of input data while ensuring that the dynamics associated with subsequent inputs remain unaffected by the preceding ones.

A pivotal distinction from the RSN model is the localization of information storage. In the $\mathbb{C}$-RSN, the information acquired during the evaluation process is concentrated within a minimal portion of the network, specifically within a single neuron. This concentration of information processing and storage into a small network segment allows the rest of the network to remain unencumbered, ready to engage in other evaluation processes. This approach not only enhances the efficiency of the network but also mirrors the specialization and division of labor observed in biological neural networks. The $\mathbb{C}$-RSN model, thus, stands as a more refined and realistic emulation of the dynamic, complex nature of biological cognition and processing \cite{huebner_schulkin_2022}.

The paper is structured as follows: In Sec. \ref{section_NNasMap}, we elaborate on the architecture of the Neural Network and delve into the intricacies of the learning process. Sec. \ref{crsn_sec_2} is dedicated to describing the forward evolution of the map underlying the $\mathbb{C}$-RSN. Our numerical results, specifically pertaining to the MNIST dataset, are presented in Sec. \ref{sec::res}. The manuscript concludes with Sec. \ref{sec::conc}, where we summarize our results and discuss potential avenues for future research.

\section{\crsn $\,\,$Neural Network architecture }
\label{section_NNasMap}

In this section, we conceptualize a neural network as a discrete map, thereby establishing the mathematical underpinnings of the \crsn. We represent the neural activity by a vector $ \vec{x} \in \mathbb{R}^N $. Simultaneously, the network's structure is modeled by a weighted adjacency matrix, $ \mathbf{W} \in \mathbb{C}^{N \times N}$, characterizing a fully connected neural network of order $N$. It is pertinent to recall from graph theory that the order of a graph refers to its number of nodes, while the size denotes the number of edges.

The architecture includes two neuron types: \textit{linear} and \textit{non-linear}. With linear we mean that on such neurons an activation function acts linearly, while with non-linear we mean that on such neurons an activation function acts non-linearly. To this end, we categorize the neuron sets as $ \mathcal{NL}=\{1,2,\dots,L-1, L\}$ (with cardinality $|\mathcal{NL}|=L$) and $ \mathcal{L}=\{L,L+1,\dots,N-1, N\} $ (with cardinality $|\mathcal{L}|=N-L$), corresponding to the \textit{non-linear} and \textit{linear} segments of the network, respectively.

The neural network, in its entirety, is characterized by a singular activation function $ \tilde{f} $, which assumes a non-linear form for $1 \leq i \leq L $ and transitions to a linear form for indices beyond $ L $. For the non-linearity within $ \tilde{f} $, we opt for a sigmoidal structure, specifically a $\tanh$ function. This design choice is elucidated through the following mathematical expression:

\begin{equation}
    \Tilde{f}(z_l) = \begin{cases}
    \tanh{(z_l)} \quad \text{if} \quad l \leq L, \\
    z_l \quad \text{if} \quad l > L,
    \end{cases}
    \label{non_lin}
\end{equation}
where $\vec{z}\in \mathbb{R}^N$ is a generic vector. This formulation captures the nuanced interplay between linear and non-linear dynamics within the neural network.

In our model, we postulate that the matrix $ \mathbf{W} $, belonging to the complex space $ \mathbb{C}^{N \times N} $, is derived through its spectral decomposition:

\begin{equation}
    \mathbf{W} = \Phi\Lambda\Phi^{-1}
\end{equation}

Here, $\Lambda$ is a diagonal matrix composed of eigenvalues $\lambda_l$, and $ \Phi $ encompasses the corresponding eigenvectors. Essentially, $ \Phi $ represents the basis for this decomposition. The choice of dealing with the above decomposition of the coupling matrix follows the spectral approach to machine learning discussed in \cite{giambagli2021machine, buffoni2022spectral, chicchi2021training,chicchi2023recurrent}.
Our model's uniqueness is further accentuated in the organization of the eigenvalues $ \lambda_l $:

\begin{equation}
    \lambda_l = \begin{cases}
    e^{2\pi i /T_l} \quad\;\; \text{if} \quad l<M, \\
    \lambda_l \in {\rm I\!R} \quad \text{where} \quad |\lambda_l|<1\quad \text{if} \quad l > M.
    \end{cases}
    \label{crsn_eq_eigval}
\end{equation}

This formulation entails that the first $ M $, with $M \leq L$, eigenvalues are predetermined to be complex values. Their magnitudes are set to equal $ 1 $, with a phase contingent on the parameter $ T_l $, termed the \textit{period}. This specification imbues the model with a rhythmic, cyclical dynamism, reflecting in the oscillatory behavior of these eigenvalues. Conversely, the second set of eigenvalues (to be trained upon optimization) is constrained to have magnitudes less than $1$, ensuring a damping effect that stabilizes the network over time. It is important to note that, despite the use of complex phases, this approach is completely different from the Complex-Valued Neural Networks in \cite{aizenberg2011complex}, as our system involves a dynamical evolution instead of static network.

\begin{figure}[!ht]
	\centering
	\includegraphics[width=1.0\textwidth]{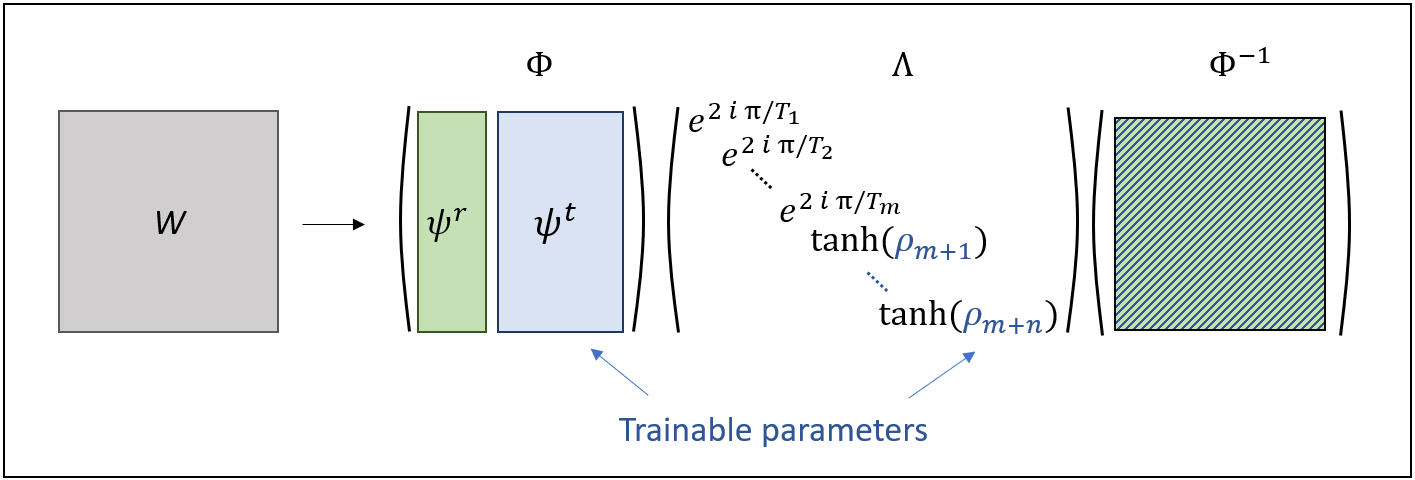}
	\caption{\footnotesize Spectral decomposition used to describe the linear transformation. The trainable parameters are highlighted in blue and contained in the set $\psi^t$. The fixed parameters are highlighted in green and contained in the set $\psi^r$. The third matrix is the inverse matrix of the basis $\Phi$ and depend to both trainable elements and fixed elements of the matrix $\Phi$. }
	\label{fig_matrix}
\end{figure}

To provide a more intuitive grasp of this concept, Fig. \ref{fig_matrix} in our paper offers a succinct visual depiction of the spectral decomposition as applied in our study. This diagrammatic representation aids in comprehending how the eigenvalues are spatially and numerically orchestrated within the matrix, highlighting the dichotomy nature of their arrangement and the distinct roles they play in the network's functionality.

To clarify, the columns of the matrix $\Phi$ correspond to the eigenvectors  $\vec{\psi}^{(k)}$ of the decomposition, where $k$ ranges from $1$ to $N$. Specifically, the first $M$ eigenvectors, denoted as $m=1,\dots,M $, are fixed and primarily localized within the linear part of the network. In Fig. \ref{fig_matrix}, this particular subset of $ M $ eigenvectors is identified by the label $ \psi^{r} $. The remaining $ N-M $ eigenvectors, encapsulated collectively under the set $ \psi^{t} $, are distinct from the first $ M $ and have different characteristics or localization within the network. 

Specifically, when $m<M$, each component of an eigenvector $\vec{\psi}^{(m)}$ is set to zero, with the exception of the $N$-th and the $(N-m)$-th components, which are set to one. In other words, for a given eigenvector $\vec{\psi}^{(m)}$, only two components are set to one: the last component and the $(N-m)$-th component, while the others are set to zero. For example, for the first three values of $m$, with $N$ and $M$ fixed, one should have:

\begin{equation}
    \vec{\psi}^{(m=1)} = 
  \begin{pmatrix}
    0 \\
    \vdots\\
    0\\
    0\\
    0\\
    1\\
    1
  \end{pmatrix}
  \hspace{4mm}
    \vec{\psi}^{(m=2)} = 
  \begin{pmatrix}
    0 \\
    \vdots\\
    0\\
    0\\
    1\\
    0\\
    1
  \end{pmatrix}
  \hspace{4mm}
    \vec{\psi}^{(m=3)} = 
  \begin{pmatrix}
    0 \\
    \vdots\\
    0\\
    1\\
    0\\
    0\\
    1
  \end{pmatrix}.
  \label{eig_form}
\end{equation}

These eigenvectors, i.e the ones in $\psi^r$, and the respective eigenvalues, are fixed and, therefore, not learnable parameters  during the training process. The ones in $\psi^t$, with the respective eigenvalues of the form:
\begin{equation}
    \lambda_l = \tanh(\rho_l) \hspace{10mm} l = M+1, \hdots, N, \hspace{5mm} \rho_l\in \mathbb{R}.
\end{equation}
are, instead, parameters that can be learned during the training process. The functional choice of the trainable eigenvectors derives from the observation that  eigenvalues magnitude must be satisfied during every single step of the training procedure. In conclusion, the Complex Recurrent Spectral Network ($\mathbb{C}$-RSN) is defined as the discrete map:

\begin{equation}
    \vec{x}_{t+1}=\tilde{f}(\Phi\Lambda\Phi^{-1} \vec{x}_t).
\label{map}
\end{equation}

The above system will be trained by employing a suitably defined loss that pivots on the real part of the time dependent  signal produced at the exit node. We will elaborate  further on this point in the following section.


\section{Forward evolution of the map}
\label{crsn_sec_2}

Here, we focus on the forward evolution of the map within our model, i.e, equation \eqref{map}. Let's begin by defining  $\vec{x}_0$ as the initial condition for our map. After $t$ iterations of the discrete map, we obtain a vector $ \vec{x}_t $.

For the purposes of this analysis, we will solely focus on the linear dynamics, by deliberately neglecting the contributions that stem from the imposed (and spatially localized) non-linearities. This assumption simplifies our understanding of the system's behavior, without losing in generality. Under linear dynamics and given the specific constraints on the eigenvalues described in the previous section, we observe that the long-term behavior of the system, i.e.,  its asymptotic dynamics, is confined to a subspace. This subspace is defined by the first $ M $ eigenvectors of the system. In principle, non-linearities could hold the potential to disrupt the system's convergence towards its expected asymptotic manifold — the subspace spanned by the eigenvectors linked to the fixed (and complex) eigenvalues.  Interestingly, our investigations reveal that such deviations from the expected behavior — which we might term \textit{divergent behaviours}  somewhat loosely — are rare even for a system that did not undergo training and that has been  initialized at random.  Therefore, in our further discussions, we will can safely assume that the dynamics of our system steadily progress, upon training, towards the final subspace defined by the fixed eigenvectors.


Let us consider a scenario where, after $t^*$ iterations, the dynamics of the map \eqref{map} become confined within the subspace formed by the first $M$ eigenvectors. This situation can be mathematically expressed as follows:

\begin{equation}
    \vec{x}_{t^*} = \sum_{m=1}^{M} \alpha_m \vec{\psi}^{(m)}. 
\end{equation}

Here, $ \vec{x}_{t^*} $ represents the state of the system after $t^*$ iterations. The expression on the right-hand side is a linear combination of the first $M$ eigenvectors $\vec{\psi}^{(m)}$, where each eigenvector is scaled by a coefficient  $\alpha_m$. This representation encapsulates the essence of our system's dynamics being dominated by these specific eigenvectors as time progresses. In other words, we are expressing $\vec{x}_{t^*}$ into a basis composed by first $M$ eigenvectors.

This formulation aligns with our earlier discussion that the system, influenced by its inherent constraints and the learning process, naturally gravitates towards this particular subspace. The coefficients $\alpha_m$ in the equation are indicative of the extent to which each of the first $ M $ eigenvectors influences the state of the system at the iteration $t^*$.

From (\ref{eig_form}) and  (\ref{non_lin}), the dynamics on the subspace is linear and the activity remains confined on it. In particular after a new application of the map \eqref{map}, the activity becomes:
\begin{equation}
    \vec{x}_{t^*+1} = \sum_{m=1}^{M} \alpha_m e^{2\pi i /T_m}  \vec{\psi}^{(m)}.
\label{crsn_eq_linear_ater_it}
\end{equation}
Making more iterations, the above equation transforms into:
\begin{equation}
    \vec{x}_{t^*+t} = \sum_{m=1}^{M} \alpha_m (e^{2\pi i/T_m})^t  \vec{\psi}^{(m)} =  \sum_{m=1}^{M} \alpha_m e^{2\pi i t/T_m}  \vec{\psi}^{(m)}.
    \label{cpx_act}
\end{equation}

Next, we turn our focus to the coefficients $\alpha_m $ in the system's dynamics. These coefficients are typically complex numbers, a result of the complex eigenvalues. Moreover, they are functions of the training parameters, which means their values are influenced by these parameters.

Depending on how a specific trainable parameter is set, the system can exhibit a variety of evolutionary patterns. These patterns, in turn, influence the coefficients $\alpha_m$, which represent the activity in the context of the chosen reference basis. Essentially, the evolution of the system under different training parameters leaves its trace in these coefficients.

To clearly illustrate this relationship, we express the coefficients  $\alpha_m$ as functions that depend on a vector of trainable parameters, denoted as $\vec{\beta}$. This approach helps to highlight how changes in the training parameters directly impact the coefficients, thereby affecting the system's overall behavior. The coefficients  $\alpha_m$ can, therefore, be expressed as follows:

\begin{equation}
    \alpha_m(\vec{\beta}) = r(m,\vec{x}_0,\vec{\beta}) + \mathrm{i} c(m,\vec{x}_0,\vec{\beta}),
\end{equation}

where $r(m,\vec{x}_0,\vec{\beta})$ and $c(m,\vec{x}_0,\vec{\beta})$ are two real scalar functions where the dependence on the initial condition $\vec{x}_0$  has been made explicit. Under this setting, the equation (\ref{cpx_act}) becomes:

\begin{equation}
\begin{split}
     \vec{x}_{t^*+t} &= \sum_{m=1}^{M} \alpha_m(\vec{\beta}) e^{2\pi \mathrm{i}  t/T_m}  \vec{\psi}^{(m)} = \sum_{m=1}^{M} (r(m,\vec{x}_0,\vec{\beta}) + \mathrm{i}  c(m,\vec{x}_0,\vec{\beta}) e^{2\pi \mathrm{i}  t/T_m} \vec{\psi}^{(m)} = \\
     &=\sum_{m=1}^{M} \big(r(m,\vec{x}_0,\vec{\beta}) + \mathrm{i}  c(m,\vec{x}_0,\vec{\beta})\big) \bigg(\cos\bigg(\frac{2\pi t}{T_m}\bigg) + \mathrm{i}  \sin\bigg(\frac{2\pi t}{T_m}\bigg)\bigg)  \vec{\psi}^{(m)}.
\end{split}
\end{equation}

From the form of eigenvalues in \eqref{eig_form}, 
and recalling that $(\vec{\psi}^{(m)})^N=1$ $\forall m=1,...,M$, we can explicitly compute the $N$-th component of $\vec{x}_{t^*+t}$:
\begin{equation}
\begin{split}
     \big(\vec{x}_{t^*+t}\big)_N &=\sum_{m=1}^{M} \bigg( r(m,\vec{x}_0,\vec{\beta})\cos\bigg(\frac{2\pi t}{T_m}\bigg) - c(m,\vec{x}_0,\vec{\beta})\sin\bigg(\frac{2\pi t}{T_m}\bigg) \bigg) + \\
     & \hspace{13mm}+ \mathrm{i} \bigg( c(m, x_0,\vec{\beta})\cos\bigg(\frac{2\pi t}{T_m}\bigg) + r(m,\vec{x}_0,\vec{\beta})\sin\bigg(\frac{2\pi t}{T_m}\bigg) \bigg),
\end{split}
\end{equation}
and taking the real part of the above equation, we end up with:
\begin{equation}
    \mathcal{R}(t,\vec{x}_0,\vec{\beta})=\operatorname{Re}{((\vec{x}_{t^*+t})_N )}= \sum_{m=1}^{M} \bigg( r(m,\vec{x}_0,\vec{\beta})\cos\bigg(\frac{2\pi t}{T_m}\bigg) - c(m,\vec{x}_0,\vec{\beta})\sin\bigg(\frac{2\pi t}{T_m}\bigg) \bigg).
\label{eq_crsn_realpart}
\end{equation}

This latter equation, represented by \eqref{eq_crsn_realpart}, plays a pivotal role in defining the loss function for our model. The focus on the real part, $\mathcal{R}(t,\vec{x}_0,\vec{\beta}) $, is not arbitrary but is rooted in the intrinsic characteristics of the system and the objectives of our training process.

In the context of our neural network model, the real part of the complex state vector, $ \vec{x}_{t^*+t} $, encapsulates critical information about the system's dynamics. By integrating the real part into the loss function, we align the training process with the goal of optimizing the network's performance based on tangible, observable outcomes. As we will detail in the following sections, the formulation of the loss function leveraging $\mathcal{R}(t,\vec{x}_0,\vec{\beta}) $ is instrumental in guiding the network towards desirable dynamics, reflecting our model's underlying principles and objectives.

Expanding upon this foundation, we apply our model to classification tasks. Our dataset $ \mathcal{D} $ consists of pairs $ (\vec{x}_0, \hat{y})^{\eta} $ for $ \eta=1, \ldots, |\mathcal{D}| $. Here, $ \vec{x}_0 $ serves as both the input vector and the initial condition for our discrete map, while $ \hat{y} $ represents the label, ranging from 1 to $C $, corresponding to different classes.

For classification purposes, we define $ C $ discrete time functions, $ f_C(t) $, each associated with a distinct class. The classification challenge involves training the network to minimize a loss function, formulated as:

\begin{equation}
    \mathcal{L} = \sum_{\eta\in\mathcal{D}}\mathcal{L}(\vec{x}_0^{\hspace{1mm}\eta},\vec{\beta})
\end{equation}

where

\begin{equation}
    \mathcal{L}(\vec{x}_0^{\hspace{1mm}\eta}, \vec{\beta}) = \sum_{t=0}^T (\mathcal{R}(t,\vec{x}_0^{\hspace{1mm}\eta},\vec{\beta}) - f_{\hat{y}^{\eta}}(t))^2.
\end{equation}

This loss function hinges on the real part of the network's output, $ \mathcal{R}(t,\vec{x}_0,\vec{\beta})$, at time $t$, with the aim of minimizing it through the adjustment of $\vec{\beta}$. The network is thus trained to align the real part of its output with the target time function for each class, over a time span $ T $, starting from the input $\vec{x}_0 $.

In the following subsections, we will delve into the results obtained from applying this model to the renowned MNIST dataset \cite{lecun1998mnist}. The MNIST dataset serves as an excellent platform to demonstrate our model's capability in classifying different classes, based on their temporal dynamics.

\section{Results}
\label{sec::res}
The MNIST dataset, an acronym for \textit{Modified National Institute of Standards and Technology}, is a renowned collection of handwritten digits. It comprises $60000$ training images and $10000$ testing images, each of size $28\times28$ pixels, making it a fundamental resource for training and testing in machine learning and image recognition domains. Over the years, it has emerged as a primary benchmark and has become the de facto standard for research in image classification. The dataset is organized into ten classes, each corresponding to the digits $0$ through $9$. Every image in the dataset is associated with a label that indicates the class of the image, i.e., the digit it represents.

In our study, we trained a $\mathbb{C}$-RSN (Complex Recurrent Spectral Network) of size $N=1000$. The non-linearity, as defined in (\ref{non_lin}), is applied to the first $L=800$ neurons. We set the number of fixed eigenvectors and eigenvalues at $M=5$. The model was trained to replicate the corresponding target discrete time function within a $T=20$ step time window, during which the loss is computed. Prior to this window, the network is allowed to evolve for $ t'=10 $ time steps, starting from an initial condition shaped by the selected image.

Each image from the MNIST dataset is normalized to ensure the pixel values range between $0$ and $1$. This normalized data is then input to a selected subset of nodes in the network, particularly those incorporating non-linear processing units. During the initial ten steps, the network executes the classification task by diverging the dynamics that result from initial conditions belonging to different classes into their respective final states.

\begin{figure}[H]
	\centering
	\includegraphics[width=0.8\textwidth]{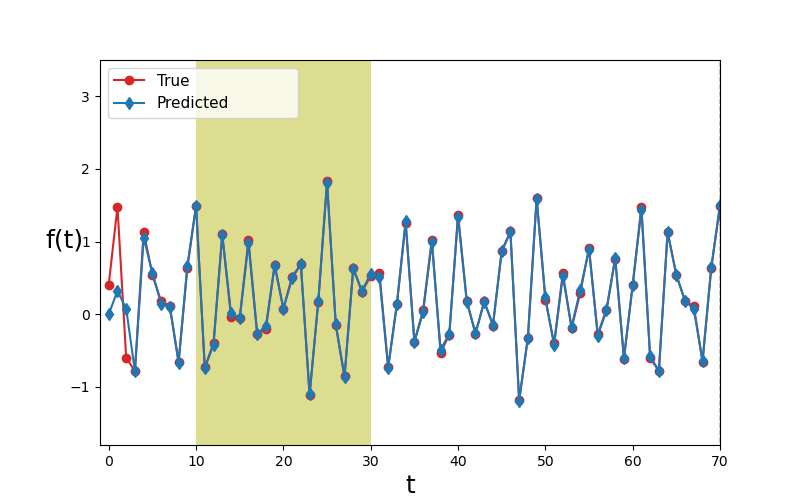}
	\caption{\footnotesize Comparison of the predicted activity in the last neuron (blue line) against the target temporal function (red line) for an input corresponding to the class $1$. The shaded yellow region indicates the time window where the loss is computed.}
	\label{CRSN_fig2}
\end{figure}

Continuing this analysis, Figure \ref{CRSN_fig2} showcases the network's performance. The discrete time function $f(t)$ output by the last neuron is depicted in blue, while the red line represents the target discrete time function for the input class. Remarkably, the two curves closely align, indicating that the model has successfully learned to reproduce the target function. Notably, this alignment persists even beyond the time window where the loss is calculated, as highlighted in yellow in the figure. Furthermore, the convergence of the curves begins even before this designated time window. These observations are consistent with findings from the RSN model presented in \cite{chicchi2023recurrent}, particularly the sustained proximity of the curves across successive time steps, which is a direct consequence of the stability of the final subspace.

To quantify the model's effectiveness, we assess its accuracy by determining, for each input, which of the potential target functions (based on L$^2$ distance) is closest to the function observed in the last neuron. Applying this method to the MNIST dataset, our model achieved an impressive accuracy of $0.9784$ on the test set. This level of accuracy is quite remarkable, especially considering that a Multi-Layer Perceptron (MLP) with a ReLu activation function, a well-established approach in the field, achieves an accuracy of around $0.9820$. The close proximity of our model's accuracy to that of the MLP underscores its efficacy. It demonstrates that our model not only achieves high accuracy but also is comparable to, and nearly matches, the performance of a conventional MLP in this context. Such results highlight the potential of our model as a robust alternative for image classification tasks, especially in scenarios where the nuanced dynamics captured by our model could offer additional benefits.

\subsection{The Role of the Basis in Signal Reconstruction}

Equation (\ref{eq_crsn_realpart}) elucidates that the signal observed in the last neuron is a linear combination of sinusoids, each defined by a distinct period $ T_m $. It is important to note that each of the fixed eigenvectors in our model is sparse, with only two non-zero elements—one consistently at position $ N$, and the other at a unique position corresponding to the eigenvector in question. The signal at these non-zero positions is, by design, a pure sinusoidal function, the period of which mirrors the $ T_m$ value specified in the definition of its associated eigenvalue. Consequently, the amplitude of these sinusoidal waves provides an indirect metric for assessing the prominence of each mode within the time-resolved patterns observed at node $N$.

\begin{figure}[H]
	\centering
	\includegraphics[width=0.6\textwidth]{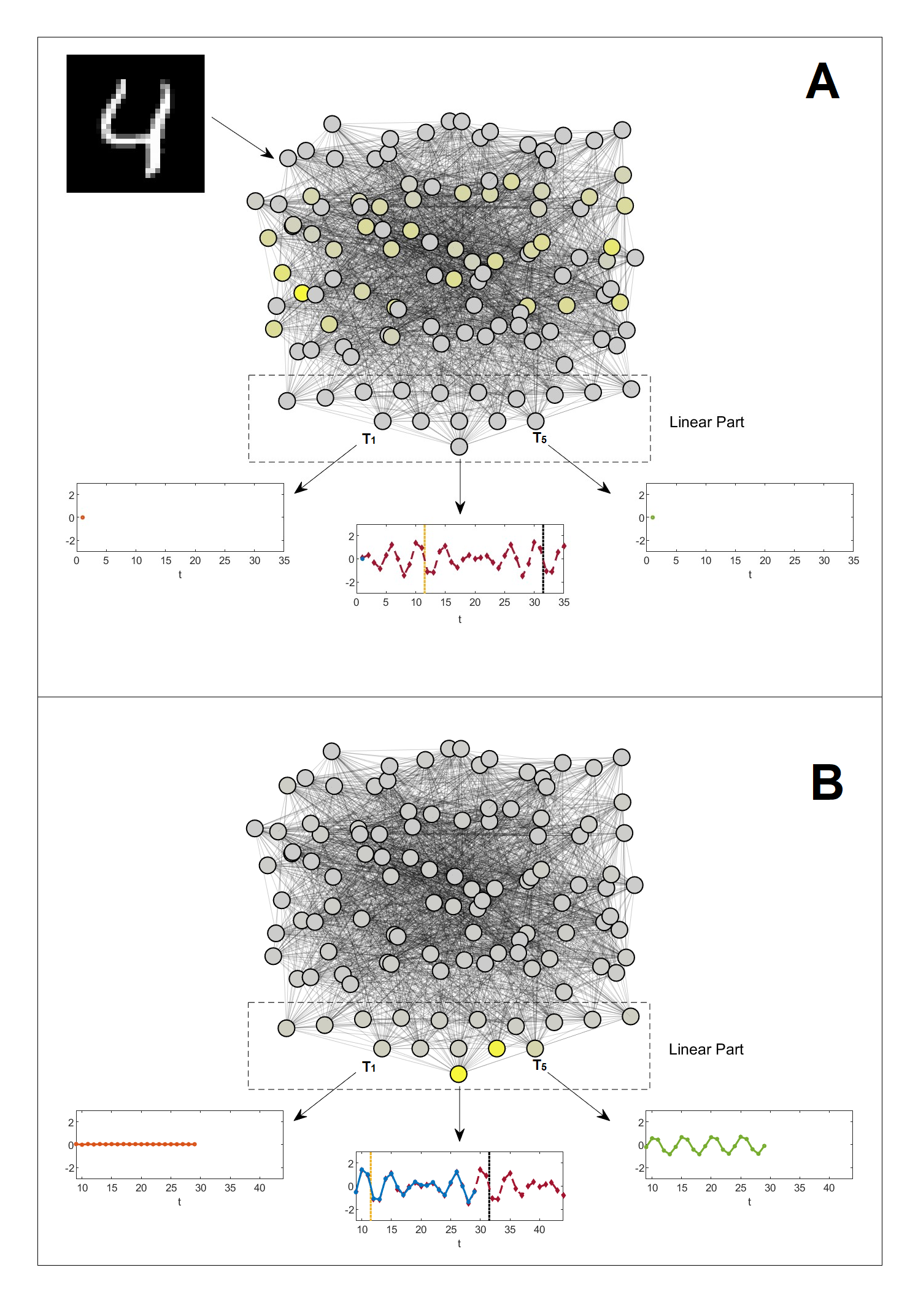}
	\caption{\footnotesize Visual representation of the \crsn $\,$network at two distinct points in its dynamic evolution. For clarity, this schematic only illustrates a subset of the network's nodes: individual nodes represent groups, except for the last node and those corresponding to the five fixed eigenvectors. The activity of the last neuron is traced by a blue line in the lower sub-panel, and the target temporal function is overlaid in red within the same sub-panel. The non-zero entries of the eigenvectors that constitute the attracting manifold are depicted as a linear horizontal array positioned just above the last neuron's activity display in the illustration. The two smaller sub-panels, to the right and left, highlight the activity in two of these nodes, specifically those characterized by periods $T_1 = \infty$ and $T_5 = 5$. Panel \textbf{A} captures the initial condition where the input data stimulates activity in the non-linear portion of the network, leaving the linear segment dormant. Panel \textbf{B}, however, presents a snapshot taken during the loss computation window, where activity has transitioned entirely to the linear part of the network. At this juncture, the signal in the last neuron aligns with the target function. The selected neurons display sinusoidal activities, the amplitudes of which correspond to the coefficients of the target function when decomposed into the Fourier basis.}
	\label{CRSN_fig3}
\end{figure}

The parameters $T_m$, introduced in (\ref{crsn_eq_eigval}), effectively filter elements from the Fourier basis, crafting a bespoke basis that the network employs to construct the signal in the last neuron. This process is displayed in Figure (\ref{CRSN_fig3}), which presents a visual representation of the network during two distinct dynamic phases: the onset of the activity and a subsequent moment within the finite window where the loss is computed.

Panel \textbf{A} of Figure (\ref{CRSN_fig3}) captures the network at the initial condition, showcasing activity initiation within the non-linear segment. As the dynamics progress, this activity transitions towards the linear portion of the network. By the time we observe Panel \textbf{B}, the network has iterated sufficiently for the activity to be fully contained within the linear section. Here, the real part of the activity in the last neuron is seen to closely resemble the target time function. Meanwhile, the neurons associated with the fixed eigenvectors display the anticipated sinusoidal behavior.

This visualization not only confirms the theoretical underpinnings of our model but also provides concrete evidence of the network’s capability to adapt and channel the activity through its architecture, culminating in the accurate reproduction of the desired signal.

\subsection{Multiple evaluations}
$\mathbb{C}$-RSN networks possess the capability to sequentially process multiple sets of input data while ensuring that the dynamics associated with subsequent inputs remain unaffected by the preceding ones. This independent processing is achievable, provided that each new data input is introduced after a duration sufficient to allow the system's dynamics from the initial data to converge into the stable subspace. We elucidate in this subsection that, under this precondition, the individual dynamics of each dataset evolve in isolation. 

To demonstrate the capability of $\mathbb{C}$-RSN networks to handle successive inputs, we decompose the activity vector $\vec{x}_t$ into two distinct components: one that corresponds to the elements within the linear segment of the network, and another that pertains to the non-linear segment. This decomposition is represented mathematically as follows:

\begin{equation}
    \vec{x}_t = \vec{x}^{\hspace{1mm}non-linear}_t + \vec{x}^{\hspace{1mm}linear}_t.
    \label{crsn_eq_split_activity}
\end{equation}
In this decomposition, the first $ L $ components of $\vec{x}^{\hspace{1mm}non-linear}_t$ are non-zero, whereas the remaining $ N-L$ components are zero. Conversely, for $\vec{x}^{\hspace{1mm}linear}_t$, the first $ L $ components are zero, and the subsequent $ N-L $ components are non-zero. This separation allows to write down the action of the activation function  $\tilde{f}$ as $ \tilde{f}(\vec{x}^{\hspace{1mm}non-linear}_t + \vec{x}^{\hspace{1mm}linear}_t) = \vec{x}^{\hspace{1mm}linear}_t + \tanh(\vec{x}^{\hspace{1mm}non-linear}_t)$. This action demonstrates the nuanced interplay between the linear and non-linear components of the network: while the linear section remains unaffected, the non-linear segment is transformed by the hyperbolic tangent function, which introduces the non-linearity essential for the network's complex behavior.

Let's now consider a time $t$ much greater than $\bar{t}$, where $\bar{t}$ represents the final time step at which the loss is evaluated. At such a time, the activity vector $\vec{x}_t$, originating from the initial condition $\vec{x}_0$, will have converged to the subspace spanned by the fixed eigenvectors. Mathematically, this is expressed as $\vec{x}_t=\sum_{m=1}^{M} \alpha_m \vec{\psi}^{(m)} $, and it can be understood that $ \vec{x}_t $ is equivalent to $ \vec{x}^{\hspace{1mm}linear}_t$, signifying that it resides solely within the linear sector of the network.
 Moreover, eq. (\ref{crsn_eq_linear_ater_it}) tells us that also $\vec{x}_{t+1}$ is confined in the same subspace and in particular:
\begin{equation}
    \vec{x}_{t+1} = \Phi\Lambda\Phi^{-1} \vec{x}_t = \vec{x}^{\hspace{1mm}linear}_{t+1}.
\end{equation}

Let's consider the introduction of an additional activity vector, $\vec{x'}_t$, which is superimposed onto the existing vector $\vec{x}_t$. The resultant neuronal activity within the network is thus captured by the composite vector $\vec{s}_t$, which is the sum of $\vec{x}_t$ and $\vec{x'}_t$, i.e, $\vec{s}_t = \vec{x}_t + \vec{x'}_t$. Performing now a new iteration on $\vec{s}_t$ one obtains:

\begin{equation}
\begin{split}
        \vec{s}_{t+1} &=  \tilde{f}(\Phi\Lambda\Phi^{-1} (\vec{x}_t + \vec{x'}_t)) 
        =\\
        &=\tilde{f}(\Phi\Lambda\Phi^{-1} \vec{x}_t + \Phi\Lambda\Phi^{-1}\vec{x'}_t)=\\
        &=\vec{x}^{\hspace{1mm}linear}_{t+1} + \tilde{f}(\Phi\Lambda\Phi^{-1}\vec{x'}_t) =\\
        &=\vec{x}_{t+1} + \vec{x'}_{t+1}.
\end{split}
\end{equation}

So, the evolution of the vectors $\vec{x}_t$ and $\vec{x'}_t $ occurs independently. This independence stems from the observation that the processing of input data results in a network state which develops within a subspace. Notably, this subspace comprises only those neurons that are part of the network’s linear segment.

The above property, therefore,  allows us to evaluate and classify different inputs sequentially by working with a trained \crsn $\,$ on a specific dataset. Indeed, after the network has completed processing a first input and reached a stable state, it can then receive and independently classify a second input.  The overall final state of the network is captured in the real part of the last neuron’s activity over time.  The time-dependent activity will be a linear superposition of the functions representing the classes of the two inputs, with a phase shift determined by the time interval between the introduction of these inputs. Mathematically, if $ t_1 $ and $ t_2$ represent the times when the inputs are introduced, the real part of the activity in the last neuron can be described as:

\begin{equation}
    \mathcal{R}(t) = f_{C_1}(t) + f_{C_2}(t+\gamma) \;\;\;\;\; \text{for } t >> t_1, t_2,
    \label{crsn_eq_linearcomb}
\end{equation}

where $ f_{C_1} $ and $f_{C_2} $ are the target functions corresponding to the classes of the inputs, and $ \gamma= t_2- t_1$ ($t_2> t_1$) is the time difference between the two inputs, i.e., $\vec{x}_0^{\eta=1}$ and $\vec{x}_0^{\eta=2}$. For the sake of simplicity, we have omitted from $\mathcal{R}(t) $ the dependence on the initial state and the vector $\vec{\beta} $, which represents the training parameters.

\begin{figure}[H]
	\centering
	\includegraphics[width=1.0\textwidth]{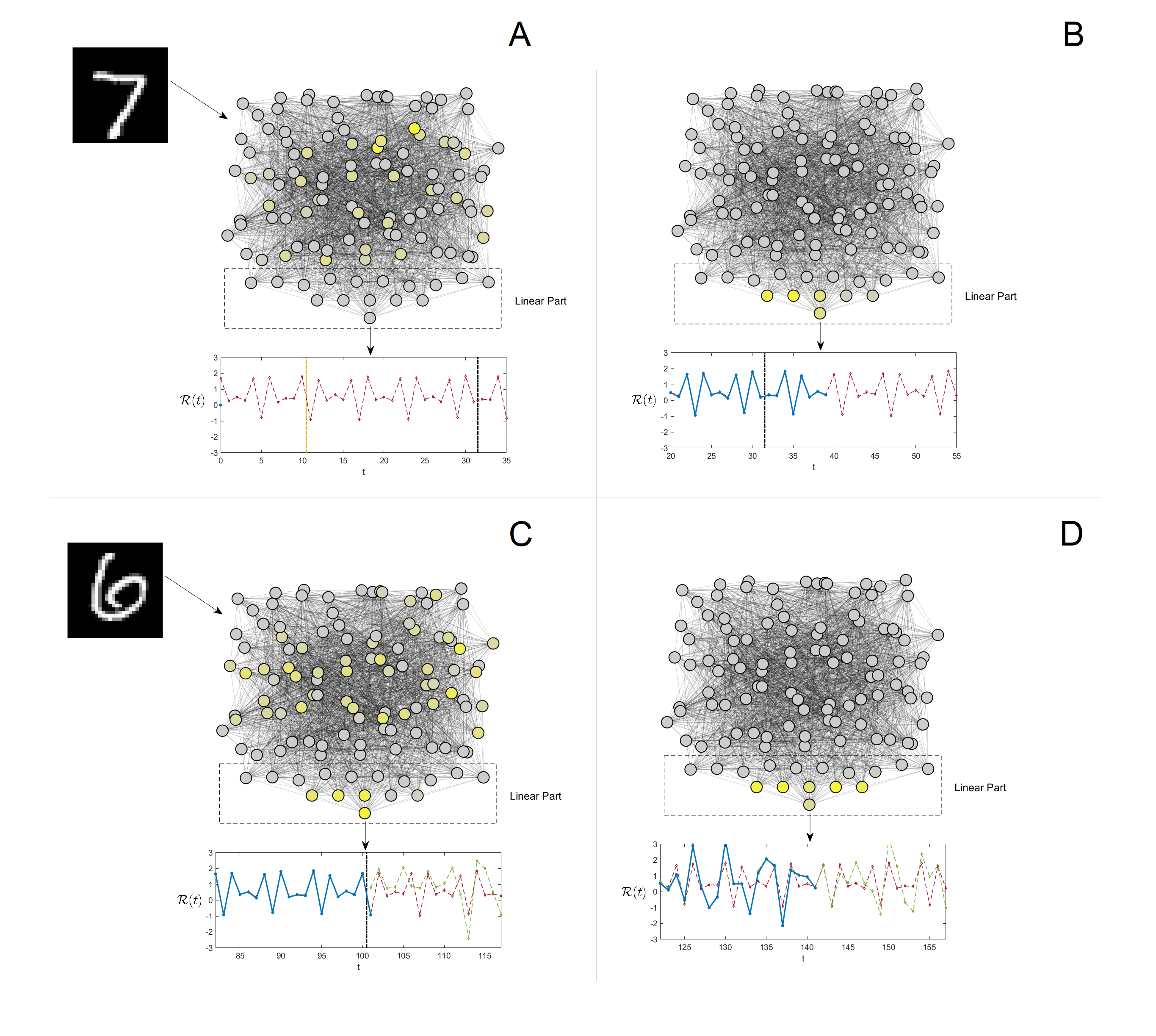}
	\caption{\footnotesize Four key stages in the evolution of a $\mathbb{C}$-RSN network trained on the MNIST dataset. Panel \textbf{A} captures the initial moment of the first data input. Panel \textbf{B} depicts a later stage where the network's dynamics have stabilized, with the sub-panel illustrating the alignment of the real part of the last neuron's activity (blue line) with the target function (red line). Panel \textbf{C} shows the introduction of a second input into the network, without removing the residual activity from the first input. Finally, Panel \textbf{D} demonstrates how the combined dynamics – from both the residual and the new input – converge back towards the linear part of the network. Notably, the wave pattern at the last neuron, corresponding to the eigenvector entries, is a linear combination of the two target functions. Intriguingly, this pattern also preserves information about the time gap between the introductions of the two inputs. }
	\label{CRSN_fig_multi}
\end{figure}

An example of sequential evaluation using the MNIST dataset is depicted in Figure (\ref{CRSN_fig_multi}), which presents four distinct phases of the network's evolution. Panel \textbf{A}, at  $t=0$, shows the initial input being introduced into the non-linear part of the network. As per the dynamics outlined in section (\ref{crsn_sec_2}), the network's activity evolves and, after a few steps (specifically at  $t=39$), it converges within the linear segment. This convergence is evident in Panel \textbf{B}, where the last neuron displays a temporal activity pattern resembling the target function. 

At $t=100$, illustrated in Panel \textbf{C}, a new input is fed into the network. This triggers the network dynamics to consolidate within the linear part once again, and the activity in the last neuron begins to reflect the linear superposition of the two target functions, in accordance with equation (\ref{crsn_eq_linearcomb}). Finally, panel \textbf{D} captures a later stage where the combined dynamics – influenced by both the residual activity from the first input and the new input – settle within the linear part. Notably, the wave pattern at the last neuron, the entry point of the eigenvectors, is shaped by a linear combination of the two target functions. This pattern intriguingly retains information about the time interval between the successive introductions of the two images.

Employing a $\mathbb{C}$-RSN model trained for individual classification, we can effectively process multiple inputs in sequence. The network's output enables the identification of both input classes and the time gap between their introductions. This ability, demonstrated by the time function in the last neuron, is a natural feature of the $\mathbb{C}$-RSN model and doesn’t require any additional modifications during training.

\section{Conclusion}
\label{sec::conc}
Artificial neural network models have historically drawn inspiration from biological insights into human brain functioning. However, these models markedly diverge from contemporary neuroscience models and experimental findings. One notable distinction is that brain activity is dynamically evolving, whereas traditional neural networks often lack this dynamic aspect.

In this manuscript we have introduced an advanced version of the RSN, termed the Complex Recurrent Network $(\mathbb{C}$-RSN), as an extension of the model presented in \cite{chicchi2023recurrent}. The $\mathbb{C}$-RSN model confines non-linearity to specific nodes, rather than allowing it to diminish over time. This design ensures that the model's characteristics remain consistent, avoiding the temporal convergence towards linearity seen in the RSN. 

Like the RSN, the $\mathbb{C}$-RSN utilizes a spectral decomposition to define neuron interactions, with a subset of eigenvectors and eigenvalues as trainable parameters. These eigenvalues are constrained to have magnitudes less than one, while the remaining eigenvectors and eigenvalues are fixed and unaltered during training. Notably, the fixed eigenvalues are complex values with a modulus of one, each correlating to a specific frequency. This structure ensures that once network dynamics enter the subspace defined by these eigenvectors, they remain confined there. Moreover, these eigenvectors converge at the last neuron, allowing for the construction of a signal as a weighted sum of sinusoidal functions corresponding to the frequencies of the fixed eigenvalues.

In this context, we formulated a classification problem where the network learns to reproduce distinct temporal activities for different input classes on the last neuron. Tested with the MNIST dataset, the $\mathbb{C}$-RSN model demonstrated exceptional accuracy. Its classification capability remains consistent over time as the system stabilizes. Once trained, the model can sequentially classify multiple inputs. The final activity observed in the last neuron not only reveals the classes of the various inputs but also the temporal intervals between their introductions, showcasing the model's advanced capacity for handling complex classification tasks.

The introduction of the $\mathbb{C}$-RSN model opens up a plethora of research avenues, particularly in fields where dynamic and complex neural network behaviors are essential. One immediate area of exploration could be in the realm of time-series analysis, where the model's ability to handle sequential inputs and maintain dynamic states could offer novel insights. This could be particularly beneficial in financial forecasting, weather prediction, or even in analyzing biological sequences.

Another promising area is in the field of neuroscience, where the $\mathbb{C}$-RSN model's bio-inspired design could help in understanding brain-like neural processing. Researchers could explore how this model simulates certain cognitive functions or neural responses, potentially offering a new perspective on neural computation in the brain.

These possibilities, along with others yet to be discovered, will be the subject of detailed investigation in forthcoming manuscripts. These future studies will aim to not only explore the full potential of the $\mathbb{C}$-RSN model but also to address the challenges and opportunities it presents in advancing the field of artificial neural networks.